%% file: iclr2025_conference.tex
\documentclass{article} 
\usepackage{iclr_fmwild,times}

\input{math_commands.tex}

\usepackage{hyperref}
\usepackage{url}
\usepackage{graphicx}
\usepackage{subcaption}
\usepackage{booktabs}

\title{Narrowing Class-Wise Robustness Gaps \\ in Adversarial Training}

\author{Fatemeh Amerehi, Patrick Healy 
\\
Computer Science and Information Systems\\
University of Limerick\\
Limerick, Ireland \\
\texttt{Fatemeh.Amerehi@ul.ie} \\
}

\iclrfinalcopy 
\begin{document}

\maketitle

\begin{abstract}

Efforts to address declining accuracy as a result of data shifts often involve various data-augmentation strategies. Adversarial training is one such method, designed to improve robustness to worst-case distribution shifts caused by adversarial examples. While this method can improve robustness, it may also hinder generalization to clean examples and exacerbate performance imbalances across different classes. This paper explores the impact of adversarial training on both overall and class-specific performance, as well as its spill-over effects. We observe that enhanced labeling during training boosts adversarial robustness by 53.50\% and mitigates class imbalances by 5.73\%, leading to improved accuracy in both clean and adversarial settings compared to standard adversarial training.
\end{abstract}

\section{Introduction}

Adversarial examples are inputs to machine learning models that have been intentionally crafted to cause incorrect model predictions. These examples are typically made by introducing slight perturbations that distort model outputs while maintaining a high degree of similarity~\cite{goodfellow2014explaining, papernot2016transferability, tramer2017space, madry2017towards}.

Formally, given an image \(x_i\) with its corresponding label \(y_i\), and a model \(f\) that correctly classifies it, i.e., \(f(x_i) = y_i\), an adversarial example \(x_i^{\prime}\) is defined as an image that satisfies two conditions: first, it causes the classifier to misclassify the image, such that \(f(x_i) \neq f(x_i^{\prime})\); and second, it remains visually similar to the original image \(x_i\)~\cite{carlini2019evaluating, engstrom2019exploring, wang2019improving}. Typically, similarity between the two images is measured using an \(\ell_{p}\)-norm, meaning that \(x_i^{\prime} = x_i + \delta\) is considered a valid adversarial example if and only if \(\lVert \delta \lVert_p \le \varepsilon\), where \(\varepsilon\) is a small constant, and \(p \in [0, \infty]\). Under these similarity constraints, adversarial examples $x_i^{\prime}$ are often crafted as in Eq.~\ref{eq:adversarial}, to maximize the loss of the model when processing the sample \(x_i\).

\begin{equation} \label{eq:adversarial}
\underset{\lVert \delta \rVert_p \le \varepsilon}{\max} \mathcal{L}(f(x_i + \delta), y_i).
\end{equation}

The set of adversarial data points generated by this maximization implicitly defines a distribution of adversarial examples~\cite{goodfellow2019research}. The Fast Gradient Sign Method (FGSM)~\cite{goodfellow2014explaining} approximates the above maximization, and generates adversarial examples by backpropagating the gradient of the loss to the input data to compute $\nabla_{x_i} \mathcal{L}(x_i, y_i)$.
This gives direction in which the loss function increases the most with respect to small changes in the input data $x_i$. 
It then moves the data in this direction (i.e., $\text{sign}\nabla_{x_i} \mathcal{L}(x_i, y_i)$) that maximizes the loss for $x_i$, as shown in Eq.~\ref{eq:adv_fgsm}: 

\begin{equation} \label{eq:adv_fgsm}
x_i^{\prime} = x_i + \varepsilon \cdot \text{sign}(\nabla_{x_i} \mathcal{L}(x_i, y_i)),
\end{equation}

where the added perturbation is scaled down by $\varepsilon$ to maintain similarity distance. 
FGSM~\cite{goodfellow2014explaining} is designed for fast generation rather than optimality. Basic Iterative Method~\cite{kurakin2016adversarial} and Projected Gradient Descent (PGD)~\cite{madry2017towards} are iterative extensions of FGSM, designed to generate stronger adversarial examples. These examples typically viewed as the worst-case form of distributional shift, where even minor perturbations can lead to significant misclassifications~\cite{rice2021robustness}. 

To address distribution shifts, data augmentation techniques are commonly used to improve model performance in image classification by increasing data variety and reducing the gap between training and test data distributions~\cite{hendrycks2018benchmarking, xu2023comprehensive}. Likewise, to enhance model robustness against attacks, adversarial training is the most effective method, involving the augmentation of training data with adversarial examples~\cite{goodfellow2014explaining, kurakin2016adversarial, moosavi2016deepfool, madry2018towards, athalye2018obfuscated}.

While data augmentation enhances overall accuracy and robustness, its effects are often highly class-dependent. Techniques like random cropping, for instance, can introduce class imbalance—improving average test performance while significantly degrading accuracy for certain classes~\cite{kirichenko2024understanding}. Augmentations may also produce unintended spillover effects; for example, color jittering strengthens robustness to brightness and color shifts yet unexpectedly weakens robustness to pose~\cite{idrissi2022imagenet}.
Similarly, while adversarial training enhances robustness, it comes at a cost. It imposes trade-offs between robustness and accuracy~\cite{tsipras2018robustness}, as well as between in- and out-of-distribution generalization~\cite{zhang2019theoretically}, while also amplifying disparities in performance across different classes. As a result, certain classes may be disproportionately disadvantaged, affecting model fairness~\cite{benz2021robustness}. 

A data augmentation policy that fails to preserve label integrity can further disrupt class balance. When applied uniformly across all classes, augmentations may degrade label information unevenly, leading to imbalances even in originally well-balanced datasets~\cite{kim2020puzzle, Balestriero2022, islam2024diffusemix}. In this paper, we aim to mitigate the class imbalance caused by adversarial training by adjusting the labels used during the adversarial training process.

\section{Related works}

Adversarial training is a form of data augmentation that incorporates adversarial examples into the training pipeline. Typically, augmentation techniques expand training data by randomly applying transformations to promote invariance, thus encouraging models to make consistent predictions across different views of each sample~\cite{geiping2022much}. In general, augmentation techniques can be classified
into  two categories: label-preserving and label-mixing techniques.

Label-preserving augmentation uses transformations on images that preserve their semantic content. While these methods have shown improvements in generalization for some factors, they can also negatively impact others~\cite{idrissi2022imagenet}. Label-mixing approaches use convex combinations of pairs of examples and their labels to encourage linear behavior between training examples, which helps to regularize the model~\cite{zhang2018mixup}. Despite their effectiveness, these methods have been found to introduce label ambiguities due to random placements of images, resulting in misleading signals for supervision~\cite{kim2020puzzle, islam2024diffusemix}. Similar to both types, Label Augmentation~\cite{amerehi2024label} aims to maintain invariance to the class identity of images while also encouraging separation between class identity and transformation.
 It does so by assigning one-hot labels \(z_j\) to each transformation operation \(o_j\) applied during augmentations. Rather than merely augmenting the transformed data \(\tilde{x_i}\) during training, without distinguishing between labels for transformed and untransformed inputs, it augments the labels by concatenating the original input labels with the operation labels:
 
\begin{equation} 
\tilde{y}_{i} = \text{Concat}[(1-\delta)y_i, \delta z_j],
\end{equation} 
where, where \(\delta\) is a scaling factor that prevents excessive deviation of the model toward the augmented label.
The training objective is thus expressed as:

\begin{equation}
\mathcal{L}_{LA}( \tilde{y}_{i}  \tilde{p}_{i}) = -\sum_{k=1}^{K+M}  \tilde{y}_{ik} \log  \tilde{p}_{ik},
\end{equation}

where, $\tilde{p}_i$ denotes the softmax of predictions for $\tilde{x_i}$, and \(\Tilde{y}_{ik}\) is a vector of length \(K+M\), which merges the original \(K\) class labels and the \(M\) transformation labels. 
This method has been shown to improve both the clean and robust accuracy. In the following sections, we examine whether Label Augmentation can be incorporated into adversarial training to reduce its negative side effects.

\section{Experimental Setup}
\label{sec:Experimental_Setup}

We study how average and class-level performance change when concatenating labels during 10-step PGD~\cite{madry2017towards} adversarial training, with $\ell_\infty$ constraints and constraint budget $\varepsilon = 0.03$. Our evaluation focuses on robustness against common and adversarial perturbations, as well as the induced side effects of adversarial training.

\textbf{Architecture and Training Details.} 
We run all experiments on an RTX-3080 GPU with CUDA Version 12.5 using PyTorch version 2.0.1. We fine-tune the ResNet50 model with default weights on the ImageNet (IN) for 10 epochs. The training starts with a learning rate of 0.01, which decays by a factor of 0.0001 according to a cosine annealing learning rate schedule~\citep{loshchilov2016sgdr}. We optimize the models using stochastic gradient descent with a momentum of 0.9. The batch sizes for training and evaluation are set to 64. We set the scaling factor \(\delta = 0.03\) to reflect the strength of the added perturbation.  
We evaluate both average error and per-class error using the original ImageNet and the corresponding robustness benchmark datasets: IN-C~\cite{hendrycks2018benchmarking}, IN-X~\cite{idrissi2022imagenet}, and IN-ReaL~\cite{beyer2020we}.  Additionally, we assess their adversarial robustness using a 40-step PGD attack, with $\ell_\infty$ constraints and constraint budget $\varepsilon = 0.03$.

\textbf{Evaluation metrics.}
The Clean error denotes the standard classification error on uncorrupted test data. For a given corruption \(c\) within IN-C, the error at severity \(s\) is denoted as \(\text{E}_{c,s}\). The Corruption Error (\(\text{CE}_c\)) is the average error over severities: \(\text{CE}_c = \frac{1}{5} \sum_{s=1}^{5} \text{E}_{c,s}\). The mean Corruption Error (\(\text{mCE}\)) is then averaged across all 15 corruptions: \(\text{mCE} = \frac{1}{15} \sum_{c=1}^{15} \text{CE}_{c}\). This single value enables comparisons against common corruptions~\cite{hendrycks2018benchmarking}. To account for label noise, we evaluate on IN-ReaL~\cite{beyer2020we}, which provides re-assessed multi-label annotations for the ImageNet validation set. These metrics quantify a model's error in assigning incorrect labels. To explore spillover effects, we evaluate on IN-X~\cite{idrissi2022imagenet}, which includes human annotations of failure modes across 16 variation factors, such as pose, size, color, and occlusions. We compute standard classification error as well as error ratio for each factor as \(\text{E}_{f} = \frac{1 - \text{accuracy}(\text{factor})}{1 - \text{accuracy}(\text{model})}\), which measures how much a model’s errors increase for a specific variation factor relative to its overall performance~\cite{idrissi2022imagenet}.

\section{Results}
\label{sec:Results}

\textbf{Average Error.} Table~\ref{tab:error_rate_vs_std} presents error rates under three evaluation conditions. Compared to the standard model, adversarial training (Adv) improves adversarial robustness by 25.24\% but compromises clean error and corruption robustness by 54.19\% and 22.99\%, respectively. Adv$^+$ further enhances adversarial robustness by 65.23\% while weakening clean error by 27.93\% and corruption robustness by 14.42\%. Both methods involve trade-offs, but Adv$^+$ achieves a better balance, mitigating the losses in clean error by 17.03\% and corruption robustness by 6.97\% compared to Adv, while achieving a 53.50\% more improvement in adversarial robustness.

\begin{figure}
    \centering
    \begin{subfigure}{0.32\textwidth}
        \centering
        \includegraphics[width=\linewidth]{ 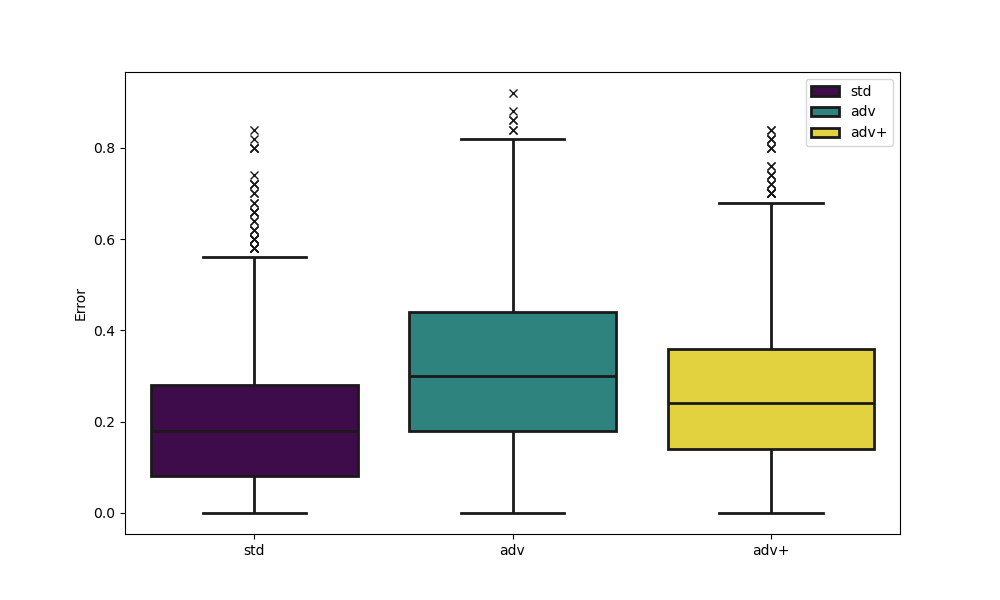}
        \caption{\footnotesize Clean ImageNet.}
        \label{fig:classwise_error_IN}
    \end{subfigure}
    \hfill
    \begin{subfigure}{0.32\textwidth}
        \centering
        \includegraphics[width=\linewidth]{ 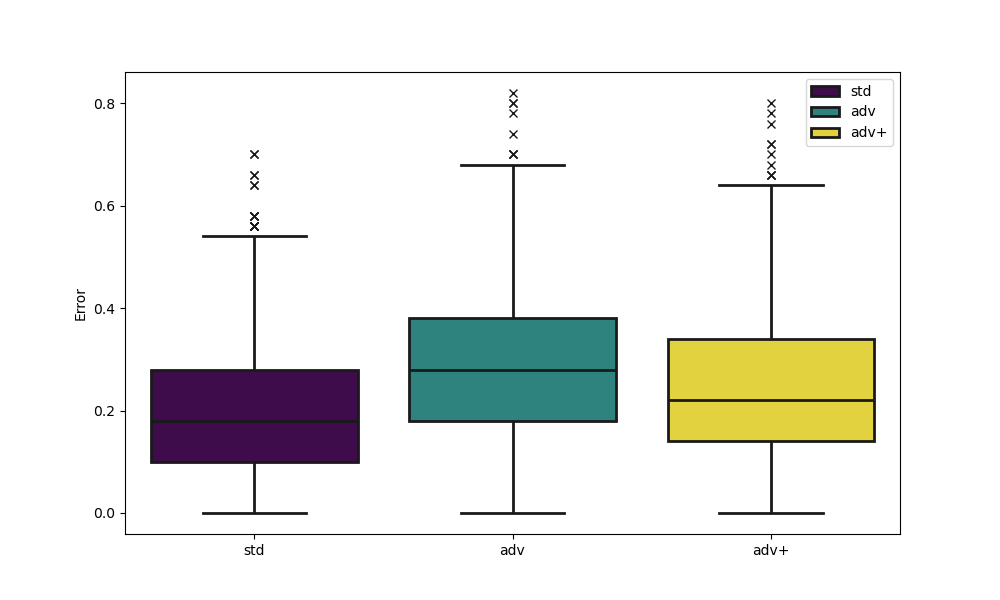}
        \caption{\footnotesize  ImageNet-ReaL.}
        \label{fig:Real_error_IN}
    \end{subfigure}
    \hfill
    \begin{subfigure}{0.32\textwidth}
        \centering
        \includegraphics[width=\linewidth]{ 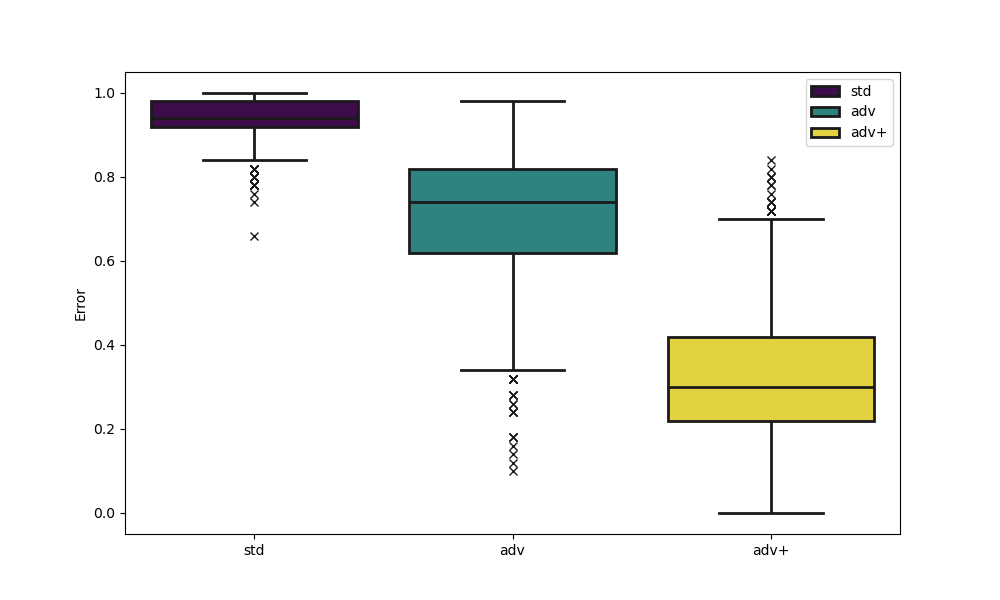}
        \caption{\footnotesize PGD-40 on ImageNet.}
        \label{fig:pgd_error_IN}
    \end{subfigure}
    \caption{Class-wise errors on different settings.}
    \label{fig:comparison_errors}
\end{figure}

\begin{table}
    \centering 
    \begin{tabular}{lccc}  
        \toprule
        \textbf{Training} & \textbf{Clean} & \textbf{mCE} & \textbf{PGD-40} \\
        \midrule
        Std     & \textbf{20.87} & \textbf{58.06} & 94.06 \\
        Adv     & 32.18 & 71.41 & 70.32 \\
        Adv$^+$  & 26.70 & 66.43 & \textbf{32.70} \\
        \bottomrule
    \end{tabular}
    \caption{ Error Rate of Adversarial Training With/Without Label Augmentation on ResNet-50}
    \label{tab:error_rate_vs_std}
\end{table}
\textbf{Class-Wise Error.} Figure \ref{fig:comparison_errors} shows the class-wise error comparessions. Both Clean ImageNet~\ref{fig:classwise_error_IN} and ImageNet-ReaL~\ref{fig:Real_error_IN} show that, compared to the standard model, adversarial methods increase the existing class-wise imbalance. However, Adv$^+$ outperforms Adv by reducing both mean error and class-wise error variability—by 5.73\% on Clean and 7.87\% on IN-ReaL settings. Under a PGD-40 attack~\ref{fig:pgd_error_IN}, the standard model exhibits the highest error and low variability, suggesting uniform vulnerability. In contrast, adversarial models exhibit similar class-wise imbalances, with Adv$^+$ reducing imbalance by 2.79\% compared to Adv while also achieving greater adversarial robustness.

\begin{table}
    \centering

    \begin{tabular}{l cc cc cc}
        \toprule
        & \multicolumn{2}{c}{Clean} & \multicolumn{2}{c}{ReaL} & \multicolumn{2}{c}{PGD-40} \\
        \cmidrule(lr){2-3} \cmidrule(lr){4-5} \cmidrule(lr){6-7}
        Method & Mean & SD & Mean & SD & Mean & SD \\
        \midrule
        std  & \textbf{20.87} & \textbf{15.65} & \textbf{19.43} & \textbf{12.78} & 94.06 & \textbf{4.71} \\
        adv  & 32.17 & 17.78 & 29.00 & 15.24 & 70.32 & 15.77 \\
        adv+ & 26.69 & 16.76 & 24.27 & 14.04 & \textbf{32.70} & 15.33 \\
        \bottomrule
    \end{tabular}
    \caption{ Mean and Standard Deviation (SD) across Different Methods}
\end{table}

\begin{figure*}
    \centering
    \includegraphics[width=0.95\textwidth]
    { 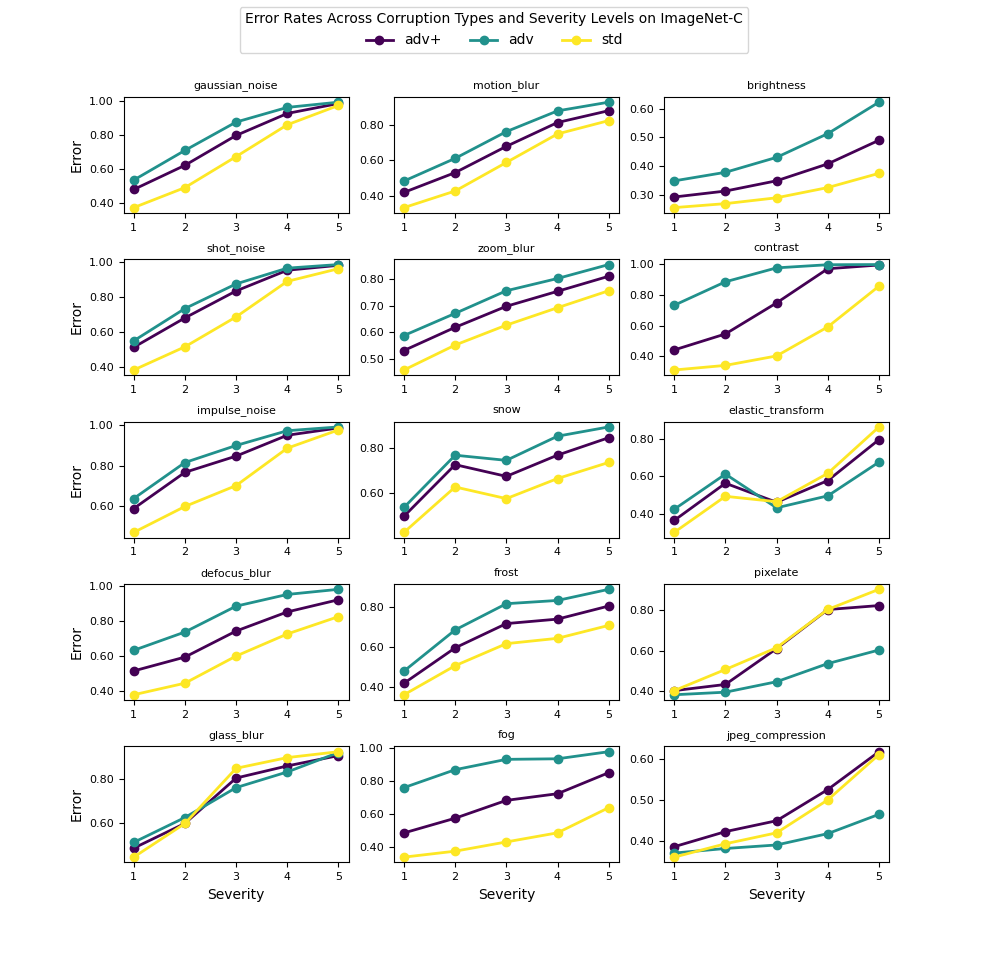}
    \caption{\footnotesize Error rates across corruption types and severity levels on ImageNet-C.}
    \label{fig:corruption_error_IN}
\end{figure*}

\begin{figure*}
    \centering
    \includegraphics[width=0.95\textwidth]{ 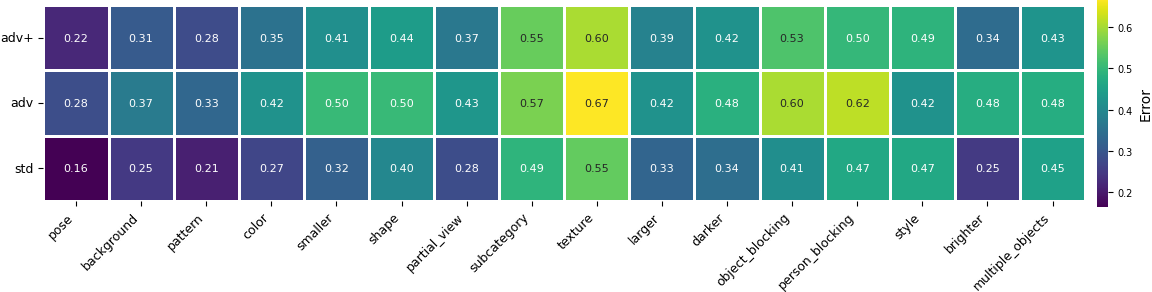}
    \caption{\footnotesize Error rate across ImageNet-X categories.}
    \label{fig:error_IN}
\end{figure*}

\begin{figure*}
    \centering
    \includegraphics[width=0.95\textwidth]{ 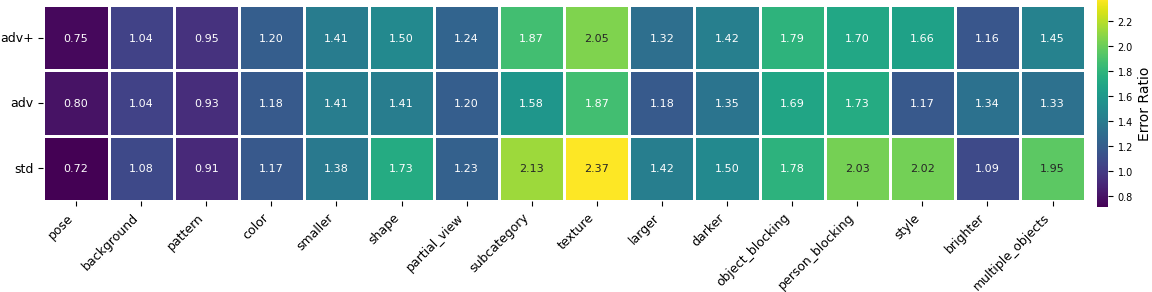}
    \caption{\footnotesize Error ratio across ImageNet-X categories.}
    \label{fig:error_ratio_IN}
\end{figure*}

\textbf{Error Rates Across Corruptions and Categories.} Figure \ref{fig:corruption_error_IN} presents error rates on ImageNet-C~\cite{hendrycks2018benchmarking} across various corruption types and severity levels. In most cases, the standard model performs better across different corruption types and severity levels. For adversarial models, Adv$^+$ consistently outperforms Adv, except in JPEG and Pixelate corruptions. Figures \ref{fig:error_IN} and \ref{fig:error_ratio_IN} show the error rates and error ratios across different IN-X categories. While the standard model achieves lower overall error, all models exhibit similar types of mistakes. However, adversarial models demonstrate improved error ratios in style and texture variations.


\section{Conclusion}
\label{sec:Conclusion}

While adversarial training helps mitigate distribution shifts from adversarial examples, it often results in reduced performance on clean samples and increased class-wise error disparities. Modifying labels during adversarial training is easy to implement, enhancing overall robustness while achieving a more favorable trade-off compared to standard adversarial training.

\subsubsection*{Acknowledgments}
This publication has emanated from research conducted with the financial support of Taighde Éireann – Research Ireland under Grant No. 18/CRT/6223.

\bibliography{iclr2025_conference}
\bibliographystyle{iclr2025_conference}

\end{document}

%% file: math_commands.tex
\usepackage{amsmath,amsfonts,bm}

\def\eqref#1{equation~\ref{#1}}

\def\1{\bm{1}}

\DeclareMathAlphabet{\mathsfit}{\encodingdefault}{\sfdefault}{m}{sl}
\SetMathAlphabet{\mathsfit}{bold}{\encodingdefault}{\sfdefault}{bx}{n}

